\documentclass[letterpaper]{article} % DO NOT CHANGE THIS
\usepackage[draft]{aaai2026}  % DO NOT CHANGE THIS
\usepackage{times}  % DO NOT CHANGE THIS
\usepackage{helvet}  % DO NOT CHANGE THIS
\usepackage{courier}  % DO NOT CHANGE THIS
\usepackage[hyphens]{url}  % DO NOT CHANGE THIS
\usepackage{graphicx} % DO NOT CHANGE THIS
\usepackage{natbib}  % DO NOT CHANGE THIS AND DO NOT ADD ANY OPTIONS TO IT
\usepackage{caption} % DO NOT CHANGE THIS AND DO NOT ADD ANY OPTIONS TO IT
\usepackage{algorithm}
\usepackage{algorithmic}
\usepackage{amsmath}
\usepackage{amsfonts}
\usepackage{booktabs}

\usepackage{newfloat}
\usepackage{listings}
\DeclareCaptionStyle{ruled}{labelfont=normalfont,labelsep=colon,strut=off} % DO NOT CHANGE THIS
\floatstyle{ruled}
\newfloat{listing}{tb}{lst}{}
\floatname{listing}{Listing}
\title{Enhancing Vision-Language Model Training with Reinforcement Learning in Synthetic Worlds for Real-World Success}
\author{George Bredis, Stanislav Dereka, Viacheslav Sinii,\\
Ruslan Rakhimov,
Daniil Gavrilov}

\makeatletter
\newcommand{\codelink}{%
  \if T\showauthors@on
    \noindent\textbf{Code: }%
    \url{https://github.com/corl-team/VL-DAC}%
  \fi
}
\makeatother

\makeatletter
\def\showauthors@on{T}   % T => show \@author; anything else => "Anonymous submission"
\def\affiliations_{T-Tech\\
Correspondence: \texttt{g.bredis@tbank.ru}}%
\makeatother

\date{July 2025}

\begin{document}

\maketitle

\begin{abstract}

Interactive multimodal agents must convert raw visual observations into coherent sequences of language-conditioned actions -- a capability that current vision-language models (VLMs) still lack. Earlier reinforcement-learning (RL) efforts could, in principle, endow VLMs with such skills, but they have seldom tested whether the learned behaviours generalize beyond their training simulators, and they depend either on brittle hyperparameter tuning or on dense-reward environments with low state variability.  
We introduce Vision-Language Decoupled Actor-Critic (VL-DAC), a lightweight, hyperparameter-free RL algorithm. VL-DAC applies PPO updates to action tokens while learning value only at the environment-step level: an arrangement, to our knowledge, not previously explored for large VLMs or LLMs. This simple decoupling removes unstable weighting terms and yields faster, more reliable convergence.  
Training a single VLM with VL-DAC in one inexpensive simulator at a time (MiniWorld, Gym-Cards, ALFWorld, or WebShop) already produces policies that generalize widely: +50\% relative on BALROG (game-centric agentic control), +5\% relative on the hardest part of VSI-Bench (spatial planning), and +2\% on VisualWebBench (web navigation), all without degrading general image understanding accuracy.
These results provide the first evidence that a simple RL algorithm can train VLMs entirely in cheap synthetic worlds while delivering measurable gains on real-image agentic, spatial-reasoning, and web-navigation benchmarks.

\end{abstract}
\codelink
\section{Introduction}

Large language models (LLMs) behave like capable single-turn agents in text-only domains, where reinforcement learning (RL) can be applied without manual annotation \citep{openai2024openaio1card,deepseekai2025deepseekr1incentivizingreasoningcapability}. Yet they still stumble when a task unfolds over many turns, revealing open problems in long-horizon reasoning and credit assignment -- the main limitation to general-purpose agency. These challenges intensify for vision-language models (VLMs) (\cite{wang2024qwen2vlenhancingvisionlanguagemodels}, \cite{chen2024internvlscalingvisionfoundation}): in addition to planning across multiple steps, a VLM must parse a constantly changing visual stream. While state-of-the-art VLMs excel at describing static images and videos, they struggle to decide \emph{what to do next} in interactive scenes \citep{chow2025physbenchbenchmarkingenhancingvisionlanguage,paglieri2024balrogbenchmarkingagenticllm}.

\begin{figure}[t]
\centering
\includegraphics[width=0.45\textwidth]{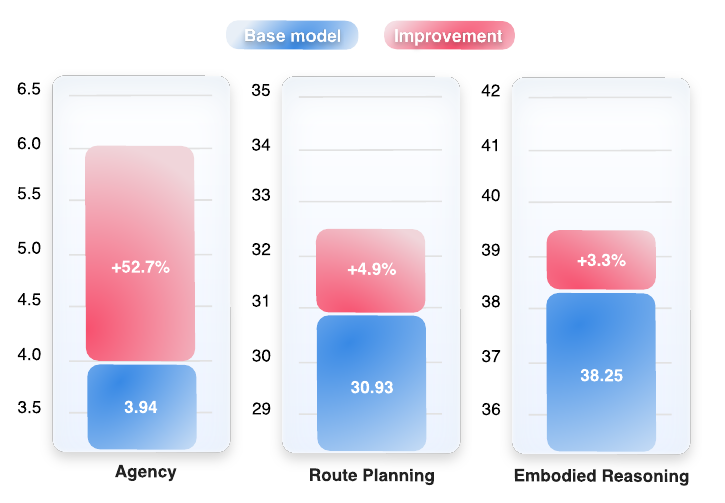}
\caption{\textbf{Real-world skill transfer after synthetic training}. Our method, VL-DAC, improves agentic control, spatial planning, and embodied reasoning on BALROG, VSI-Bench, and ERQA. It demonstrates effective transfer from synthetic environments to real-world benchmarks.}
\label{fig:teaser}
\end{figure}

Collecting genuine, step-by-step vision-language interaction data is expensive and slow; as a result, most training corpora contain only static image-text pairs, so VLMs become excellent describers but poor actors. If we want to teach a model agentic skills or an understanding of dynamic environments, we need methods and data that provide such knowledge; arguably, multi-step training in dynamic environments might be the main path toward such learning. Simulators provide a cheaper workaround, yet existing RL algorithms still stumble. RL4VLM \citep{rl4vlm} depends on a brittle coefficient that mixes "thought" and action probabilities, and slight mis-tuning makes learning diverge. LOOP \citep{putta2024agentqadvancedreasoning} aggregates reward across an entire action sequence, so its credit assignment breaks down when successive states vary greatly. ArCHer \citep{zhou2024archertraininglanguagemodel} counters variance with a learned critic, but the critic trains well only under dense rewards and a substantial off-policy buffer. Both of them are hard to maintain in long, sparse visual episodes.

\paragraph{What we find.}
Experiments in several \emph{lightweight} simulators (MiniWorld \citep{MinigridMiniworld23}, Gym-Cards, ALFWorld \citep{shridhar2021alfworldaligningtextembodied}, and WebShop \citep{yao2023webshopscalablerealworldweb}) reveal that transferable visuomotor skills emerge when two ingredients are present: (i) a simulator that is cheap enough to try many tasks, and (ii) an RL recipe that can be dropped in without delicate retuning. Training a single VLM in \emph{one} simulator at a time still lifts performance on natural-image agentic, spatial-reasoning, and web-navigation benchmarks, showing that realism and scale are \emph{less} limiting than the practicality of the learning rule. This recipe unlocks a path toward environment scaling and scenarios in which one can easily train a model in new environments and switch between them without brittle hyperparameter tuning, learning new skills off the shelf.

\paragraph{Our solution: Vision-Language Decoupled Actor-Critic (VL-DAC).}
To meet that practicality requirement, we propose VL-DAC, an RL objective that cleanly separates the learning signals:
\begin{itemize}\setlength{\itemsep}{2pt}
\item \emph{Action loss}-token-wise Proximal Policy Optimization \citep{schulman2017proximalpolicyoptimizationalgorithms}.
\item \emph{Value loss}-computed once per environment step, with gradients stopped at the VLM backbone.
\end{itemize}
This token/step split, to our knowledge unused at VLM scale, eliminates RL4VLM's brittle weighting term, avoids LOOP's sequence-level credit-assignment pitfalls, and dispenses with ArCHer's bulky replay buffer and reward requirement. The outcome is a concise, environment-agnostic algorithm that converges faster and ports across simulators with minimal fuss-exactly what is needed to push RL-trained VLMs into new domains at low cost.

\paragraph{Contributions}
\begin{itemize}
\item \textbf{Vision-Language Decoupled Actor-Critic (VL-DAC)}. We propose an RL objective that pairs token-wise PPO updates with a step-level value head whose gradients are stopped at the VLM backbone; a minimal stabilization kit (KL regularization, value warm-up, and stop-gradient) lets VL-DAC train without the fragile weighting terms or replay buffers required by earlier methods.
\item \textbf{Cheap-simulator $\rightarrow$ real-task transfer}. Training the same VLM in \emph{one} lightweight simulator at a time (MiniWorld, Gym-Cards, ALFWorld, or WebShop) already yields sizable relative gains on different benchmarks. This shows that simulator affordability and algorithm simplicity are the key ingredients for transfer.
\item \textbf{Skill-transfer study}. We provide the first systematic analysis of how simulator-acquired skills map onto agentic, spatial, and web-interaction benchmarks, and ablate each VL-DAC component to pinpoint the elements that drive stability and generalization.
\end{itemize}

Taken together, our results demonstrate that a modest algorithmic tweak, combined with low-cost simulators, suffices to unlock practical RL training for VLMs, endows them with transferable real-world competence, and opens a path toward environment scaling and large-scale learning from experience.

\section{Background}

\subsection{Vision-Language Agents in Interactive Environments}
\label{sec:mdp}

We model each episode as a finite-horizon Markov Decision Process (MDP)
$\mathcal{M} = \langle\mathcal{S},\mathcal{A},\mathcal{P},\mathcal{R},\gamma\rangle$,
where $\gamma\!\in\![0,1)$ is the discount factor.
Unlike classical RL, the \emph{state} $s_t\!\in\!\mathcal{S}$ is a tuple
$(\mathbf{x}_t,\,\mathbf{c}_t)$ consisting of an RGB image (or stack of images) $\mathbf{x}_t\!\in\!\mathbb{R}^{H\times W\times3}$ and an optional text context $\mathbf{c}_t$ (system prompt, dialogue history, etc.).

The \emph{action} $a_t\!\in\!\mathcal{A}$ is a sequence of natural-language tokens
that fully specifies the next low-level step in the environment
(e.g., \texttt{"turn\_left 15"} or \texttt{"click\_button id=OK"}).

An agent executes a trajectory
$\tau = (s_1,a_1,\dots,s_T,a_T)$
and seeks to maximize the discounted return
\[
J(\theta)
=\;
\mathbb{E}_{\tau\sim \pi_\theta}
\Bigl[
\sum_{t=1}^T \gamma^{\,t-1}\,\mathcal{R}(s_t,a_t)
\Bigr],
\]
where the \emph{policy} $\pi_\theta(a_t\!\mid\!s_t)$ is parameterized by a large vision-language model (VLM) and factorizes auto-regressively,
\[
\pi_\theta(a_t\!\mid\!s_t)
\;=\;
\textstyle\prod_{i=1}^{|a_t|}
\pi_\theta\bigl(a_t^{(i)} \mid s_t, a_t^{(<i)}\bigr).
\]
During training, we may additionally learn a state-value function
$V_\phi(s_t)=\mathbb{E}_{\tau\sim\pi_\theta}
[\sum_{k\ge 0}\gamma^{\,k}\mathcal{R}(s_{t+k},a_{t+k})]$, but the way action and value updates interact differs across methods,
as reviewed next. In VL-DAC, we retain this shared backbone but prevent value-head gradients from flowing back, thereby eliminating cross-signal interference.

\subsection{Existing RL Algorithms for Multi-Step VLMs \& LLMs}
\label{sec:related_rl}

Below, we summarize the three baselines that dominate recent work and pinpoint
the specific pain points that motivate our \emph{Vision-Language Decoupled Actor-Critic} (VL-DAC)
objective introduced in Section~\ref{sec:VL-DAC}.

\paragraph{RL4VLM \citep{rl4vlm}.}
The policy is decomposed into a ``thought’’ segment
($a^{\text{thought}}$) and an ``action’’ segment
($a^{\text{action}}$). RL4VLM multiplies token-logits of the thought span by $\lambda\!\in\![0,1]$, effectively rescaling gradient magnitudes:
\begin{align}
\label{eq:rl4vlm}
\log\pi_\theta(a_t \mid s_t)
=\;& \notag\\
=\lambda\,\log\pi_\theta(a_t^{\text{thought}}\!\mid\!s_t)&
+\log\pi_\theta\bigl(a_t^{\text{action}}\!\mid\!s_t,a_t^{\text{thought}}\bigr),
\end{align}
after which, PPO updates are applied at the \emph{step} level.
But $\lambda$ needs to be tuned for each model-environment setup.  
This makes it hard to scale the method beyond a single environment and limits environment scaling.

\paragraph{LOOP \citep{chen2025reinforcementlearninglonghorizoninteractive}.}
LOOP employs leave-one-out advantage estimation and trains an LLM in a multi-step scenario using PPO. Because it uses PPO, different policy-update levels (token, step, and trajectory) can be explored; the authors show that the best quality is achieved at the token level.  
LOO advantage estimation:
% \begin{align}
% A
%   &= \frac{K}{K-1}\Biggl(
%         R\bigl(s_{0:T},a_{0:T}\bigr)
%         -\frac{1}{K}\sum_{j=1}^{K}
%           R\bigl(s_{0:T},a_{0:T}\bigr)
%       \Biggr).
% \end{align}
\begin{equation}
\resizebox{0.9\linewidth}{!}{$
\begin{aligned}
A
  &= \frac{K}{K-1}\Biggl(
        R\bigl(s_{0:T},a_{0:T}\bigr)
        -\frac{1}{K}\sum_{j=1}^{K}
          R\bigl(s_{0:T},a_{0:T}\bigr)
      \Biggr)
\end{aligned}
$}
\end{equation}
\noindent The approach sidesteps any need for tuning token mixtures but suffers from extreme credit-assignment noise: a single bad token can wipe out the reward signal for the entire chain, making long-horizon tasks hard to learn.

\paragraph{ArCHer \citep{zhou2024archertraininglanguagemodel}.}
ArCHer trains a critic with bootstrapped one-step TD \citep{sutton1988learning} at the step level and trains the actor LM from critic feedback. Since the method is primarily designed to be off-policy, it requires a large replay buffer.  
The method works under \emph{dense} rewards, but two practical issues emerge when we want to train on-policy (e.g., when it is hard to maintain a large buffer) or have sparse rewards (due to the critic design):
\begin{itemize}
\item \textbf{Replay bottleneck}. Memory demands grow with episode length, which is acute for vision tasks where each step embeds a high-dimensional image, multiple images, or video.
\item \textbf{Reward sparsity}. When rewards arrive only at episode termination,
   the critic’s bootstrap targets become nearly constant, offering little
   learning signal.
\end{itemize}

\begin{figure}[t]
\centering
\includegraphics[width=0.45\textwidth]{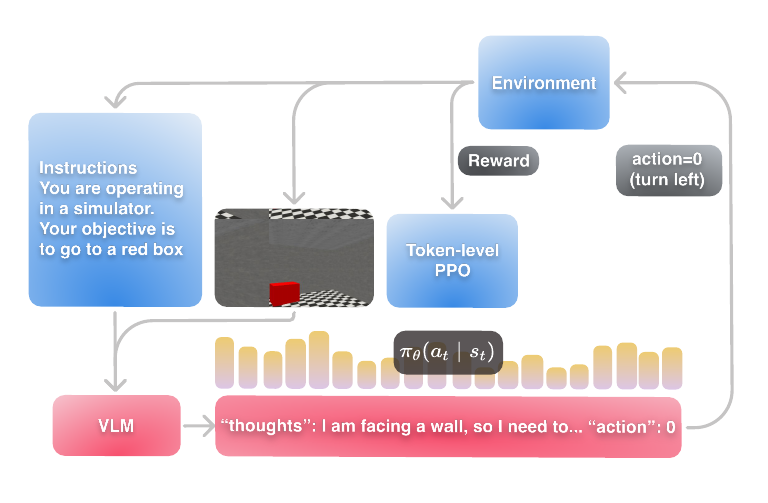}
\caption{\textbf{Vision-Language Decoupled Actor-Critic (VL-DAC) pipeline.} A vision-language model receives RGB frames and text context, predicts token-wise actions via PPO, and learns a step-level value head whose gradients are stopped at the backbone.}
\label{fig:method_overview_VL-DAC}
\end{figure}

\section{Vision-Language Decoupled Actor-Critic (VL-DAC) Training}
\label{sec:VL-DAC}

VL-DAC retains the intuitive separation between reasoning (\textbf{thought}) and behaviour (\textbf{action}) tokens advocated by RL4VLM, but removes the additional coefficient by moving the policy loss to the \textbf{token} level while keeping the value loss at the \textbf{step} level. Figure \ref{fig:method_overview_VL-DAC} presents the overall pipeline of our approach.

\paragraph{Token-level policy loss.}
Although token-wise reinforcement losses have been explored in earlier language or vision-and-language work (\cite{misra2017mappinginstructionsvisualobservations}, \cite{chen2025reinforcementlearninglonghorizoninteractive}), they have not been combined with a step-level value objective nor systematically evaluated on modern high-capacity VLMs. Our contribution is therefore to integrate this granularity in multi-step VLM scenarios with a Vision-Language Decoupled Actor-Critic design that yields greater stability without extra hyperparameters (besides the PPO \citep{schulman2017proximalpolicyoptimizationalgorithms} parameters). Let $a_t = (a_t^{1},\ldots,a_t^{|a_t|})$ denote the tokens emitted at time $t$. We apply the PPO objective independently to each token:
% \begin{align}
% &\mathcal{L}^{\text{VL-DAC}}_{\text{policy}}(\theta) = \notag\\
% &-\mathbb{E}_{\tau} \Bigl[ \tfrac{1}{|a_t|}\sum_{i=1}^{|a_t|}
% \min\Bigl( r_{t,i} A_t, \text{clip}(r_{t,i},1-\epsilon,1+\epsilon)A_t \Bigr) \Bigr],
% \label{eq:VL-DAC_tok}
% \end{align}
\begin{equation}
\resizebox{0.9\linewidth}{!}{$
\begin{aligned}
&\mathcal{L}^{\text{VL-DAC}}_{\text{policy}}(\theta) = \\
&-\mathbb{E}_{\tau} \Bigl[ \tfrac{1}{|a_t|}\sum_{i=1}^{|a_t|}
\min\Bigl( r_{t,i} A_t, \text{clip}(r_{t,i},1-\epsilon,1+\epsilon)A_t \Bigr) \Bigr],
\end{aligned}
$}
\label{eq:VL-DAC_tok}
\end{equation}
\noindent where $r_{t,i} = \pi_\theta(a_t^i\mid s_t,a_t^{<i})/\pi_{\theta_{\text{old}}}(a_t^i\mid s_t,a_t^{<i})$ and the advantage $A_t$ is still computed at the \emph{step} level using GAE \citep{schulman2018highdimensionalcontinuouscontrolusing}.

\paragraph{Step-level value loss.}
$V_\phi$ shares the backbone with $\pi_{\theta}$ but has its own MLP head. The value head predicts $V_\phi(s_t)$ once per environment step:
\begin{equation}
V_\phi(s_t)=\text{MLP}_\phi\bigl(\mathcal{F}_{\text{VLM}}(s_t)\bigr).
\end{equation}
The value loss is
$\mathcal{L}^{\text{Value}}(\phi)=\tfrac12\bigl(V_\phi(s_t)-\hat{R}_t\bigr)^2$. For step-level advantage estimation, we use GAE \citep{schulman2018highdimensionalcontinuouscontrolusing}.

\paragraph{Stabilization.}
For stabilization, we employ well-known techniques from the classical RL setup \citep{lehmann2024definitiveguidepolicygradients}, but these are currently underexplored in large language-model scenarios. We warm up $\phi$ for $n$ epochs before updating $\theta$, use \texttt{StopGrad} for the value head, and apply a per-token forward KL penalty:
\begin{equation}
\mathcal{L}^{\text{KL}}(\theta)=\mathbb{D}_{\text{KL}}\bigl(\pi_{\theta}(\cdot\mid s_t)\,\|\,\pi_{\text{old}}(\cdot\mid s_t)\bigr).
\end{equation}

\paragraph{Full objective.}
The final training loss combines the three terms:
\begin{equation}
\mathcal{L}(\theta,\phi)=\mathcal{L}^{\text{VL-DAC}}_{\text{policy}}(\theta)+\beta\,\mathcal{L}^{\text{KL}}(\theta)+\alpha\,\mathcal{L}^{\text{Value}}(\phi).
\end{equation}
We show empirically that this simple decoupling yields more stable learning curves and higher final returns than both RL4VLM \citep{rl4vlm} and LOOP \citep{chen2025reinforcementlearninglonghorizoninteractive}. We further demonstrate that simple RL training transfers the learned skills to downstream benchmarks. For the concrete prompting setup, refer to Appendix A.

\begin{figure*}[t]
\centering
\includegraphics[width=0.8\textwidth]{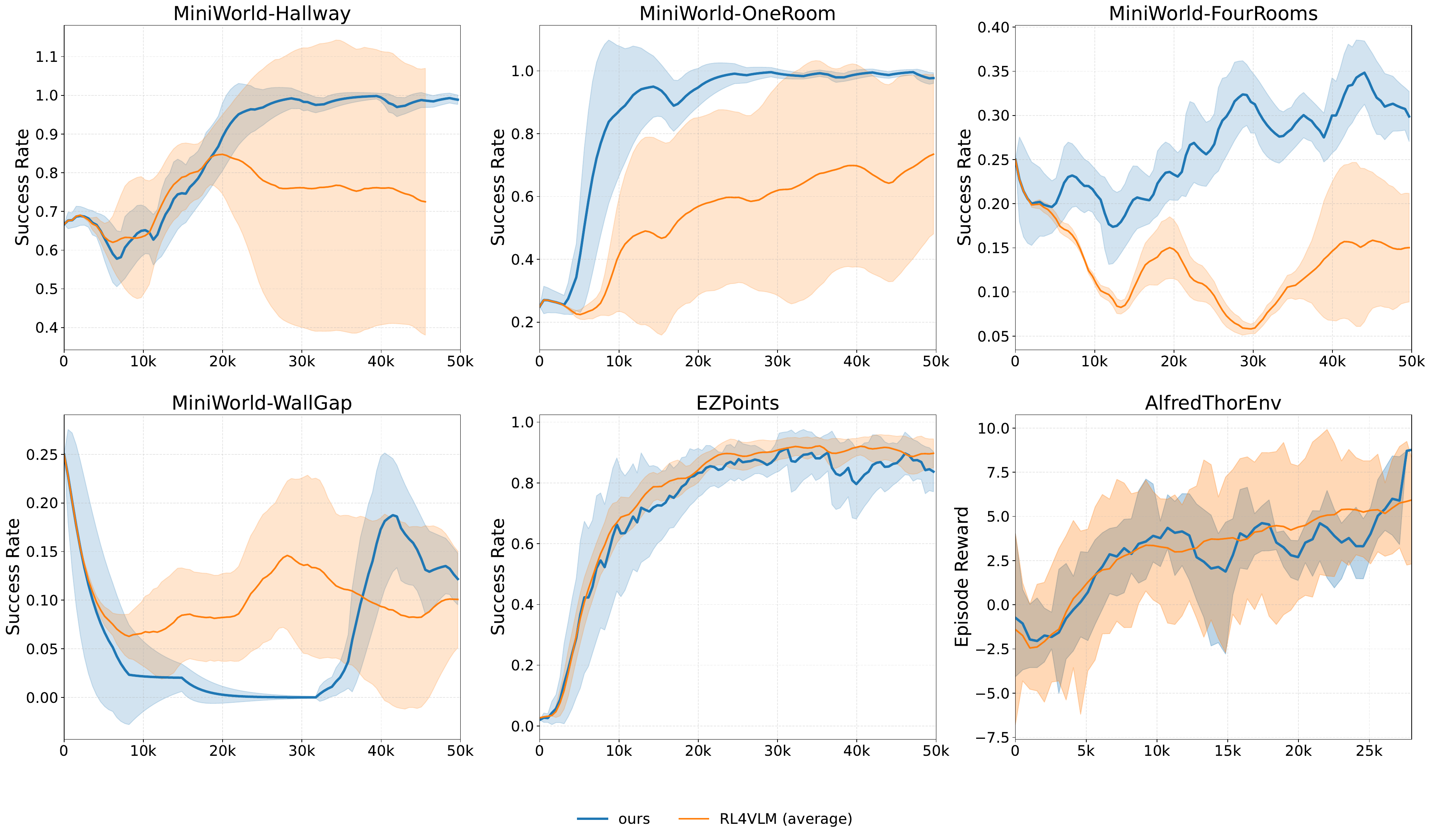}
% \caption{Episode success rates (\%) of our method and RL4VLM, averaged over multiple $\lambda$ values, on different environments. First row, left to right: MiniWorld-Hallway, MiniWorld-OneRoom, MiniWorld-FourRooms. Second row, left to right: MiniWorld-WallGap, EZPoints, ALFWorld. While the optimal choice of $\lambda$ varies significantly across environments, our method, requiring no tuning of this hyperparameter, consistently achieves strong and stable performance across all settings.}
\caption{\textbf{Episode success rates across environments}. Success rates (\%) of our method vs. RL4VLM (averaged over multiple $\lambda$ values) on six environments: MiniWorld-Hallway, OneRoom, FourRooms (top row), WallGap, EZPoints, ALFWorld (bottom row). While RL4VLM requires tuning $\lambda$ per environment, our method performs robustly without tuning.}
\label{fig:q1_stability}
\end{figure*}

\section{Experiments}
\label{sec:experiments}

Our study asks four questions:
\begin{enumerate}
\item[\textbf{Q1}] Does \textbf{VL-DAC} train more simply (in terms of hyper-parameter search) than \textbf{RL4VLM} in diverse simulators? We also explore how each stabilization tweak (KL, value warm-up, stop-gradient) contributes to performance, how brittle RL4VLM’s $\lambda$ can be (beyond the original exploration), and how our method performs in models of different sizes and architectures.
\item[\textbf{Q2}] How does VL-DAC compare with \textbf{LOOP} when long multi-step credit assignment is required?
\item[\textbf{Q3}] Do policies learned in one cheap simulator transfer skills to benchmarks, especially on agentic tasks?
\item[\textbf{Q4}] Is the method scalable to tasks that require long-term planning, such as WebShop, and how does such training contribute to the web benchmark?
\end{enumerate}
We first summarize the experimental setup, then tackle the four questions in turn. We do not include ArCHer in the main-text experiments, since it works poorly under the on-policy scenario (where the training buffer equals the replay buffer) and when rewards are sparse (due to the one-step TD bootstrap). For experiments with ArCHer, see Appendix E.

\subsection{Setup}
\label{sec:exp_setup}

\paragraph{Simulators.}
We use several lightweight environments: \textbf{MiniWorld} (four navigation tasks) for navigation and route-planning, \textbf{Gym-Cards/EZPoints} (card-selection logic) as an easy-to-check environment, \textbf{ALFWorld} (text-conditioned household tasks) for navigation, spatial reasoning, and agentic capabilities, and \textbf{WebShop} (e-commerce browsing) as a domain that requires long-term understanding and web-based planning.  
All produce RGB frames plus a textual instruction; the agent answers with free-form text that consists of thoughts and action tokens. The total response is parsed into environment actions.

\paragraph{Model and training.}
Unless noted otherwise, we finetune \texttt{Qwen2-VL-7B} \citep{wang2024qwen2vlenhancingvisionlanguagemodels} with LoRA \citep{hu2021loralowrankadaptationlarge} adapters for 25k-50k environment steps. If a table refers to the model as \texttt{base}, it corresponds to \texttt{Qwen2-VL-7B}, unless stated otherwise. For the hyperparameter grid, check Appendix B.

\paragraph{Evaluation metrics.}
\emph{Simulator success rate} (SR) is the percentage of episodes that reach the goal.  
\emph{Skill transfer} is assessed using skill-based benchmarks (and their subsets), along with a suite of captioning tasks to check for regressions. For the full evaluation setup, see Appendix C.

\paragraph{Compute budget.}
Training VL-DAC for 50k environment steps on \texttt{Qwen2-VL-7B} takes \textbf{20 GPU-hours} on a single NVIDIA H100-80GB.

\subsection{Q1. Stability: VL-DAC vs. RL4VLM}
\label{sec:q1}

\paragraph{Comparison with RL4VLM.}
Figure \ref{fig:q1_stability} plots SR over 50k steps for \textit{Hallway}, \textit{FourRooms}, \textit{OneRoom}, \textit{WallGap}, \textit{ALFWorld}, and \textit{Gym-Cards}.
Curves for RL4VLM are shown as an average of the thought-coefficient $\lambda$ values recommended by the authors; VL-DAC uses the same optimizer and other hyperparameters, with no extra tuning.  
VL-DAC reaches high SR in five of six tasks, whereas RL4VLM diverges or plateaus whenever $\lambda$ is not properly tuned. All RL4VLM experiments here use the same stabilization techniques as VL-DAC. For results without average and additional details on runs, see Appendix D.

\paragraph{Stabilization ablation.}
Figure \ref{fig:q5_ablate} shows SR on \textit{OneRoom} when we add KL regularization, value warm-up, and stop-gradient one at a time on top of RL4VLM ($\lambda{=}0.3$, the best setting for \textit{OneRoom} in our experiments).  
Each component improves convergence speed and reduces variance; all three together boost convergence, and adding VL-DAC on top further increases training stability and final quality. The illustrated standard deviation intervals were obtained with four different seeds.

\begin{table}[t]
\centering
\begin{tabular}{c|c|c}
\toprule
Model & Setup & SR \\
\midrule
Qwen2-VL-7B & RL4VLM ($\lambda=0.35$) & $0.98 \pm 0.00$ \\
Qwen2-VL-7B & RL4VLM ($\lambda=0.5$)  & $0.93 \pm 0.07$ \\
Qwen2-VL-7B & Ours                    & $0.98 \pm 0.02$ \\
Gemma3-4B   & RL4VLM ($\lambda=0.35$) & $0.55 \pm \textbf{0.38}$ \\
Gemma3-4B   & RL4VLM ($\lambda=0.5$)  & $0.82 \pm \textbf{0.14}$ \\
Gemma3-4B   & Ours                    & $0.93 \pm 0.05$ \\
\bottomrule
\end{tabular}
\caption{\textbf{RL4VLM vs.\ ours.} Evaluated on Qwen2-VL and Gemma over four seeds with varying $\lambda$. Qwen2-VL peaks at $\lambda\!=\!0.35$ in \textit{OneRoom}, while Gemma prefers $\lambda\!=\!0.5$. Our method is robust and low-variance across both, even on the harder Gemma task.}
\label{tab:q1_different_lambda_and_model_setup}
\end{table}

\paragraph{Model and $\lambda$ comparison.}
Table \ref{tab:q1_different_lambda_and_model_setup} reports RL4VLM peak SR across different $\lambda$ values and models, alongside VL-DAC’s off-the-shelf run. To produce standard deviations, we ran each model under the same setup with four different seeds. RL4VLM training with different models and $\lambda$ setups shows huge changes in both the standard deviation and the best SR, whereas our method works consistently, independently of the setup. Interestingly, for RL4VLM, the optimal $\lambda$ changes with the model, and on Gemma3-4B \citep{gemmateam2024gemmaopenmodelsbased}, RL4VLM exhibits a very large standard deviation regardless of $\lambda$, which casts doubt on its practical usability.

\textbf{Bottom line.} VL-DAC inherits the best of RL4VLM after the stabilization tweaks \emph{and} removes the hyperparameter that still limits RL4VLM in practice due to the need for tuning.

\begin{figure}[t]
\centering
\includegraphics[width=0.45\textwidth]{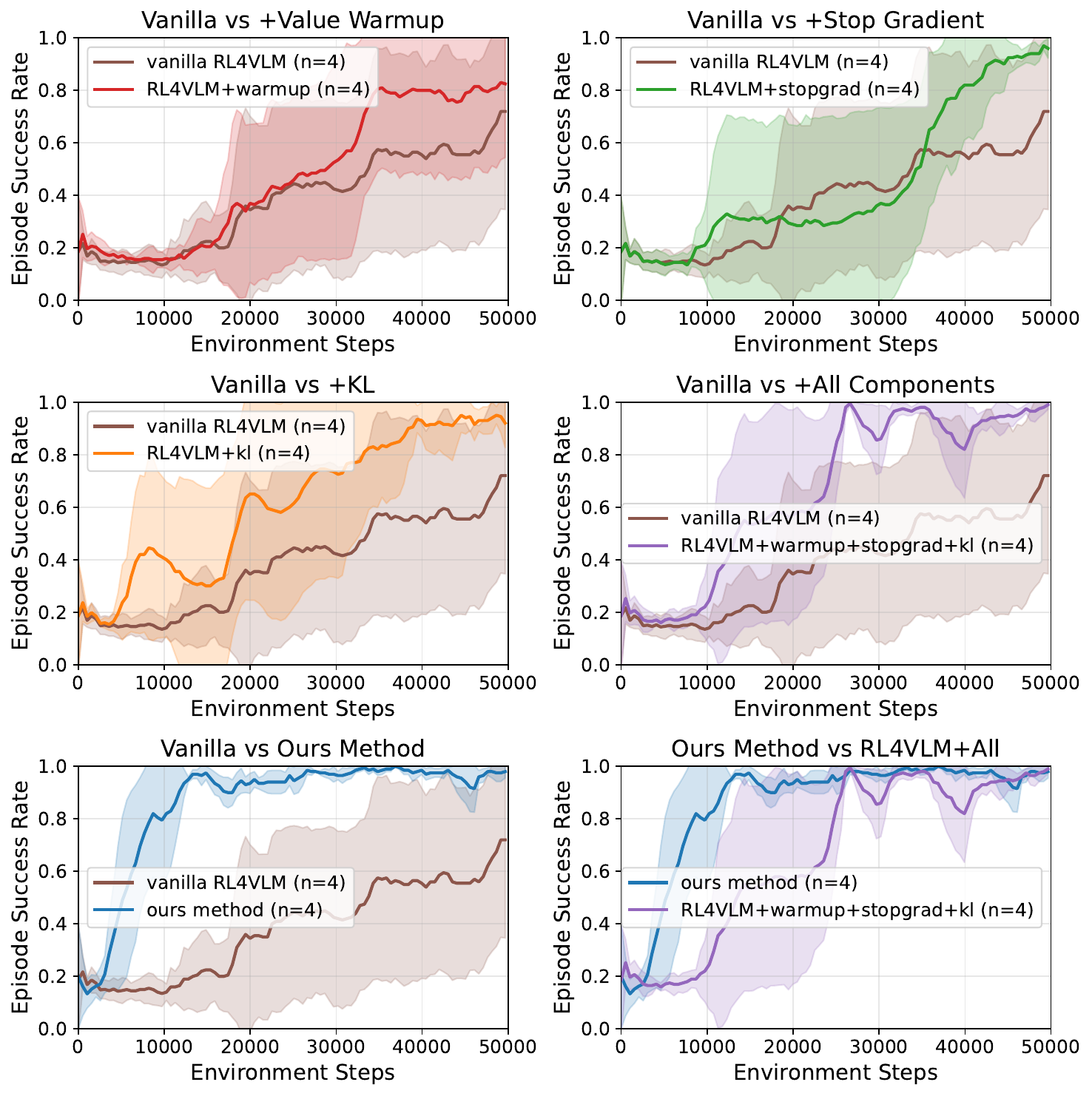}
\caption{\textbf{Ablation study of stabilization tricks.} Adding KL regularization, value warm-up, and stop-gradient cuts variance sequentially; replacing the step-level policy loss with VL-DAC's token-level objective yields the smooth ascent reported in Figure \ref{fig:q1_stability}.}
\label{fig:q5_ablate}
\end{figure}

\subsection{Q2. Long-horizon credit: VL-DAC vs. LOOP}
\label{sec:q2}

\begin{figure*}[ht!]
\centering
\includegraphics[width=0.8\textwidth]{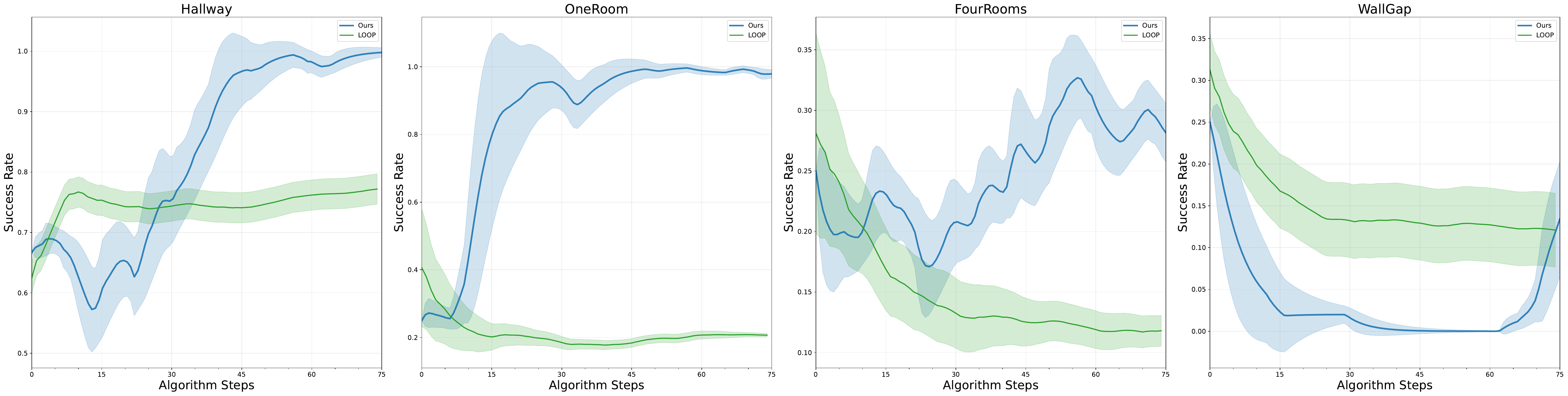}
\caption{\textbf{Long-horizon credit assignment: VL-DAC vs.\ LOOP}. On four sparse-reward MiniWorld tasks, LOOP plateaus once early successes exhaust its high-variance sequence-level gradient, whereas VL-DAC continues improving. Token-wise advantages coupled with a step-wise critic unlock sustained learning.}
\label{fig:q2_loop}
\end{figure*}

On four sparse-reward MiniWorld environments (Hallway, FourRooms, OneRoom, and WallGap), we trained LOOP \citep{chen2025reinforcementlearninglonghorizoninteractive} and VL-DAC. As Figure \ref{fig:q2_loop} shows, LOOP’s success rate plateaus after 15-30k steps, whereas VL-DAC keeps climbing. The difference stems from credit assignment: LOOP feeds the same noisy, sequence-level return to every token, while VL-DAC’s step-level critic delivers stable advantages throughout training.

In long-horizon, sparse-reward settings, sequence-level methods like LOOP stall, whereas VL-DAC’s decoupled token/step objective continues improving, yielding up to +34 pp higher success without extra tuning.

\subsection{Q3. From MiniWorld/ALFWorld to skill-based benchmark tests}
\label{sec:q3}

Tables \ref{tab:q3_balrog_transfer} and \ref{tab:q3_specific_benchmark_transfer} list downstream scores after training in \emph{one} simulator at a time.

\begin{table}[t]
\centering
\begin{tabular}{l|c|c}
\toprule
 & Base & ALFWorld-tuned \\
\midrule
$Balrog_{naive}$ & $3.21\% \pm 0.75\%$ & $\mathbf{4.19\% \pm 0.92\%}$ \\
$Balrog_{CoT}$   & $3.94\% \pm 0.98\%$ & $\mathbf{6.02\% \pm 1.19\%}$ \\
\bottomrule
\end{tabular}
\caption{\textbf{Balrog performance across prompting strategies}. RL training (notably VL-DAC) raises scores even with naive prompts, and Chain-of-Thought prompting adds a further $>$50\% boost.}
\label{tab:q3_balrog_transfer}
\end{table}

$BALROG$\citep{paglieri2024balrogbenchmarkingagenticllm} probes long-horizon agentic skills required to solve videogames, $VSI$-Bench \citep{yang2025thinkingspacemultimodallarge} subsets test spatial reasoning and planning, $ERQA$ \citep{erqa} checks spatial reasoning, $MuirBench$ \citep{wang2024muirbench} covers multi-image understanding, and $\text{VideoMME}_{\text{spatial}}$ \citep{fu2024videommefirstevercomprehensiveevaluation} checks spatial understanding.

\begin{itemize}
\item \textbf{BALROG:} +50\% relative gain in agentic success after ALFWorld training (mean ± std across four seeds), demonstrating that multi-step environments chiefly improve agentic skills.
\item \textbf{Skill-specific benchmarks:} +5 \% relative on the VSI-Bench Route-Planning task following ALFWorld training. Training in \textit{OneRoom} also yields substantial gains on VSI-Bench subsets. We extended ERQA evaluation to \textbf{naive} and \textbf{CoT} \citep{wei2023chainofthoughtpromptingelicitsreasoning} prompting and show improvements in both cases. Gains are also observed on MuirBench and the spatial subset of VideoMME. All results are single-seed due to the dataset scale.
\item \textbf{Image and video understanding:} Table \ref{tab:benchmarks} shows that the model does not lose skills on general-purpose benchmarks (\cite{zhang2024lmmsevalrealitycheckevaluation}, \cite{fu2024videommefirstevercomprehensiveevaluation}, \cite{zhang2024mmerealworldmultimodalllmchallenge}, \cite{yue2024mmmumassivemultidisciplinemultimodal}, \cite{liu2024mmbenchmultimodalmodelallaround}, \cite{ying2024mmtbenchcomprehensivemultimodalbenchmark}, \cite{hudson2019gqa}, \cite{zhao2024benchmarking}, \cite{fu2024videommefirstevercomprehensiveevaluation}, \cite{chen2024rightwayevaluatinglarge}, \cite{yan2025mirbenchllmrecognizecomplicated}),  on after training and sometimes even improves.
\end{itemize}

Also, it is important to note that earlier research indicates that supervised learning needs accurate, large-scale data curation to yield small improvements in a similar set of benchmarks without degrading performance on others.
\begin{table*}[t]
\centering
\resizebox{\textwidth}{!}{%
\begin{tabular}{l|c|c|c|c|c|c}
\toprule
 & $\text{VSI-Bench}_{\text{route plan}}$ & $\text{VSI-Bench}_{\text{relative direction}}$ & $\text{ERQA}_{\text{naive}}$ & $\text{ERQA}_{\text{CoT}}$ & MuirBench & $\text{VideoMME}_{\text{spatial}}$ \\
\midrule
Base           & 30.93 & 32.01 & 38.25 & 39.00 & 41.23 & 64.8 \\
ALFWorld-tuned & \textbf{32.47} & 31.61 & 39.00 & \textbf{39.25} & \textbf{42.58} & \textbf{66.7} \\
OneRoom-tuned  & 31.96 & \textbf{33.05} & \textbf{39.25} & 38.50 & 41.12 & \textbf{66.7} \\
\bottomrule
\end{tabular}}
\caption{\textbf{Skill-specific benchmarks}. Models trained in two different environments outperform the base model in their corresponding skill categories.}
\label{tab:q3_specific_benchmark_transfer}
\end{table*}

\begin{table}[t]
\centering
\resizebox{\columnwidth}{!}{%
\begin{tabular}{l|c|c|c|c}
\toprule
Benchmark & Base & ALFWorld & OneRoom & Hallway \\
\midrule
GQA                           & 62.02 & \textbf{62.35} & 62.06 & 62.12 \\
Mirb                          & \textbf{37.38} & 36.64 & 37.25 & 37.25 \\
$\text{MMBench}_{\text{dev}}$ & 78.86 & 78.52 & \textbf{79.04} & 78.52 \\
$\text{MME}_{\text{perception}}$ & 1681 & \textbf{1688} & 1670 & 1678 \\
MMERealWorld                  & 41.81 & 41.46 & 41.76 & \textbf{42.01} \\
$\text{MMStar}_{\text{avg}}$  & 56.53 & 57.03 & \textbf{57.51} & 57.26 \\
$\text{MMT-mi}_{\text{val}}$  & 59.90 & 60.40 & \textbf{60.66} & 60.47 \\
$\text{MMT}_{\text{val}}$     & 62.10 & 62.36 & 62.65 & \textbf{62.71} \\
VideoMME                      & 57.70 & \textbf{58.11} & 57.40 & 57.70 \\
\bottomrule
\end{tabular}}
\caption{\textbf{Benchmark gains} for Qwen2-VL finetuned on ALFWorld, MiniWorld-Hallway, and MiniWorld-OneRoom. The finetuned model surpasses its instruct baseline in temporal and spatial reasoning, multi-image/video comprehension, and embodied-AI tasks.}
\label{tab:benchmarks}
\end{table}

\subsection{Q4. A different domain: WebShop $\rightarrow$ VisualWebBench}
\label{sec:q4}

We next train in \textbf{WebShop} for only 2k steps (due to compute budget). VL-DAC lifts VisualWebBench accuracy by up to +2 pp on different subsets over the base model, showing that even short interaction budgets can improve certain skills. We also explore how models trained in other environments affect the same benchmark. Mean and std computed across 4 seeds.

\begin{table*}[ht!]
\centering
\resizebox{\textwidth}{!}{%
\begin{tabular}{l|c|c|c|c|c|c|c}
\toprule
 & web caption & webqa & heading ocr & element ocr & element ground & action prediction & action ground \\
\midrule
$base_{naive}$ & $27.81\pm0.11$ & $\mathbf{71.44\pm0.00}$ & $75.62\pm1.26$ & $82.36\pm0.00$ & $\mathbf{87.49\pm0.14}$ & $4.98\pm0.00$ & $83.50\pm0.00$ \\
$base_{cot}$   & $28.38\pm0.20$ & $61.11\pm0.11$ & $74.83\pm0.00$ & $78.75\pm0.01$ & $83.29\pm0.00$ & $6.17\pm0.21$ & $78.32\pm0.56$ \\
$WS_{naive}$   & $\mathbf{29.31\pm0.02}$ & $70.32\pm0.00$ & $\mathbf{76.34\pm0.00}$ & $\mathbf{83.49\pm0.22}$ & $87.33\pm0.14$ & $5.34\pm0.00$ & $82.52\pm0.00$ \\
$WS_{cot}$     & $29.04\pm0.12$ & $62.58\pm0.05$ & $72.66\pm0.00$ & $79.95\pm0.00$ & $84.02\pm0.00$ & $\mathbf{6.41\pm0.00}$ & $78.64\pm0.00$ \\
$OR_{naive}$   & $28.19\pm0.00$ & $70.91\pm0.00$ & $74.03\pm0.12$ & $83.31\pm0.19$ & $86.68\pm0.00$ & $3.91\pm0.00$ & $\mathbf{84.47\pm0.00}$ \\
$OR_{cot}$     & $29.21\pm0.00$ & $59.89\pm0.00$ & $74.44\pm0.34$ & $76.20\pm0.24$ & $83.78\pm0.00$ & $6.05\pm0.00$ & $78.64\pm0.00$ \\
\bottomrule
\end{tabular}}
\caption{\textbf{VisualWebBench breakdown.} A 2k-step WebShop run lifts overall accuracy; web-caption and UI-action metrics benefit most. \textit{WS} refers to WebShop, \textit{OR} to OneRoom.}
\label{tab:q4_specific_benchmark_transfer}
\end{table*}

\section{Discussion}

\subsection{From a Simple Recipe to a Two-Stage Roadmap}

Our results suggest a concise two-stage recipe for turning a vision-language model into a competent interactive agent. \textbf{Stage 1} is algorithmic: adopt a token-wise PPO objective coupled with a step-wise value head. This decoupling, realized in VL-DAC, removes thought-action mixture coefficients, replay buffers, and other brittle knobs, giving a \emph{hyper-parameter-free} learner that scales from 4 B to 7 B models without retuning. \textbf{Stage 2} is environmental: feed the learner one of several lightweight simulators that span different action semantics-navigation, manipulation, card logic, and browser interaction. Stage 1 guarantees a simple RL recipe; Stage 2 supplies the behavioural coverage necessary for real-world transfer.

\subsection{Why Simulator Diversity Matters}

Performance improvements grow with new skills. ALFWorld alone imparts agentic priors that lift BALROG success by over 50 \% relative; ALFWorld and MiniWorld inject spatial planning and reasoning that raise VSI-Bench by 5 \% relative; and WebShop injects UI-sequencing patterns that boost VisualWebBench by 2 \%. Diverse simulators enhance a wider range of skills.

\subsection{Limitations and Open Challenges}

\begin{itemize}
\item \textbf{Sparse-reward variance}. Although the critic converges even with terminal rewards, the method still struggles in hard, sparse-reward settings.
\item \textbf{Beyond screen-based tasks.} All environments studied here involve discrete interface actions on rendered images; continuous-control robotics remains untested.
\item \textbf{Single-agent assumption}. VL-DAC does not address cooperative or adversarial multi-agent settings where credit must be distributed across agents.
\item \textbf{Memory and planning}. Current models struggle to process and train in environments that require long-term abstract memory and planning (e.g., MiniWorld-WallGap). 
\item \textbf{Model scale and task demands}. Our evaluation covers 4--7B-parameter models; we have not yet assessed smaller (below 1B) or much larger (tens to hundreds of billions) models. Additionally, successful training requires models to produce strictly structured, machine-parsable outputs and to maintain coherent chain-of-thought reasoning across steps. 
\end{itemize}

\subsection{Future Directions: Scaling the Environment Spectrum}

A promising next step is to procedurally generate curricula that expand both task horizon and required skill set as model capacity grows, akin to the role of MineDojo \citep{fan2022minedojo} or Crafter \citep{hafner2022benchmarkingspectrumagentcapabilities} in open-world RL. We envision an open RL4VLM Gym where each contribution adds a \emph{small, cheap} environment rather than a single monolithic photorealistic world. Such a repository would enable systematic study of \emph{environment-set scaling laws}: how many distinct interaction types are required for an additional \(n\)\% transfer gain? Algorithmically, VL-DAC could pair with hierarchical RL, using the step-level value head to supervise sub-goal policies while token-wise PPO refines low-level text actions, or integrate memory-augmented transformers to curb variance as horizons exceed 100 steps.

\subsection{Connection to Prior Work}

\paragraph{VLM and LLM training in multi-step scenarios.}
RL4VLM \citep{rl4vlm}, LOOP \cite{chen2025reinforcementlearninglonghorizoninteractive}, ArCHer \citep{zhou2024archertraininglanguagemodel}, and some other domain-specific methods (\cite{putta2024agentqadvancedreasoning}, \cite{bai2025digiqlearningqvaluefunctions}, \cite{bai2024digirltraininginthewilddevicecontrol}) pursue long-horizon training, yet they rely on delicate mixture coefficients, sequence-level gradients with high variance, or replay buffers that collapse under sparse rewards. VL-DAC inherits the stability of PPO-based RLHF while, for the first time, demonstrating \emph{consistent transfer} across agentic, spatial, and web-interaction tasks using the \emph{same} hyperparameters.
 % Figure \ref{fig:performance_example} qualitatively illustrates this leap: after ALFWorld training, the model correctly infers object relationships in ERQA, a capability absent in the base model. 
These findings underscore that a minimal algorithmic tweak, coupled with a diversified simulator set, is sufficient to unlock practical RL training for VLMs and to endow them with real-world competence.

\paragraph{Benchmarking.}
Classical perception-centric suites such as MMBench, MME, and Video-MME are indispensable for gauging static understanding, but they lack the \emph{agentic} dimension, a capacity to decide and act under long-horizon feedback. Recent game-based evaluations like \textbf{BALROG} \citep{paglieri2024balrogbenchmarkingagenticllm} and \textbf{VideoGameBench} \citep{zhang2025videogamebenchvisionlanguagemodelscomplete} close this gap by measuring whether models can plan, execute, and adapt inside fully interactive worlds that resemble classic reinforcement-learning settings. Our study leverages both families: the perception benchmarks verify that VL-DAC training leaves core recognition intact, whereas BALROG \cite{paglieri2024balrogbenchmarkingagenticllm} exposes the gains in goal-directed control. The contrast underscores a key takeaway: \textbf{agentic evaluation is where progress now moves fastest}, and RL with brittle hyperparameters can translate simulator experience into measurable improvements on these harder benchmarks.

\paragraph{Real-task transfer.}
Generalization from synthetic practice to real-world queries has been actively explored in \emph{single-step} reasoning research (\citep{chen2025enigmatascalinglogicalreasoning}, \citep{stojanovski2025reasoninggymreasoningenvironments}). Our findings extend that evidence to the \emph{multi-step} regime: VL-DAC-trained VLMs master spatial-navigation, manipulation, and web-interaction skills in cheap simulators and then transfer them to BALROG \citep{paglieri2024balrogbenchmarkingagenticllm}, VSI-Bench \citep{yang2025thinkingspacemultimodallarge}, and VisualWebBench \citep{liu2024visualwebbenchfarmultimodalllms} with only modest domain gaps. By showing that interactive rehearsal scales beyond toy boards and text puzzles to full visual control loops, we strengthen the emerging view that \emph{procedural curricula plus lightweight RL} offer a practical path toward robust real-task competence.

\section{Conclusion}

This work demonstrates that reinforcement learning in synthetic, interactive environments is a powerful and scalable strategy for enhancing vision-language models. By moving from coupled action-and-critic optimization to decoupled (two-level) optimization and introducing stabilization techniques, we significantly improve the stability and generalization of RL-based training for VLMs. Our approach avoids brittle hyperparameter tuning while achieving competitive success rates across diverse environments. More importantly, we show that models trained in these synthetic settings generalize effectively to skill-specific and general-purpose benchmarks-outperforming strong baselines without additional supervision. These findings position RL as a viable, data-efficient alternative to traditional supervised fine-tuning, opening new directions for training embodied, multimodal agents that reason and act in complex visual domains. Future work will explore scaling to more realistic 3D worlds and integrating longer-horizon planning into vision-language training.
\bibliography{aaai2026}

\clearpage
% \ifreproStandalone
\appendix
\label{appendix:raw}

\section{Appendix A: Input example}\label{app:input}
\begin{figure}
    \centering
    \includegraphics[width=0.5\textwidth]{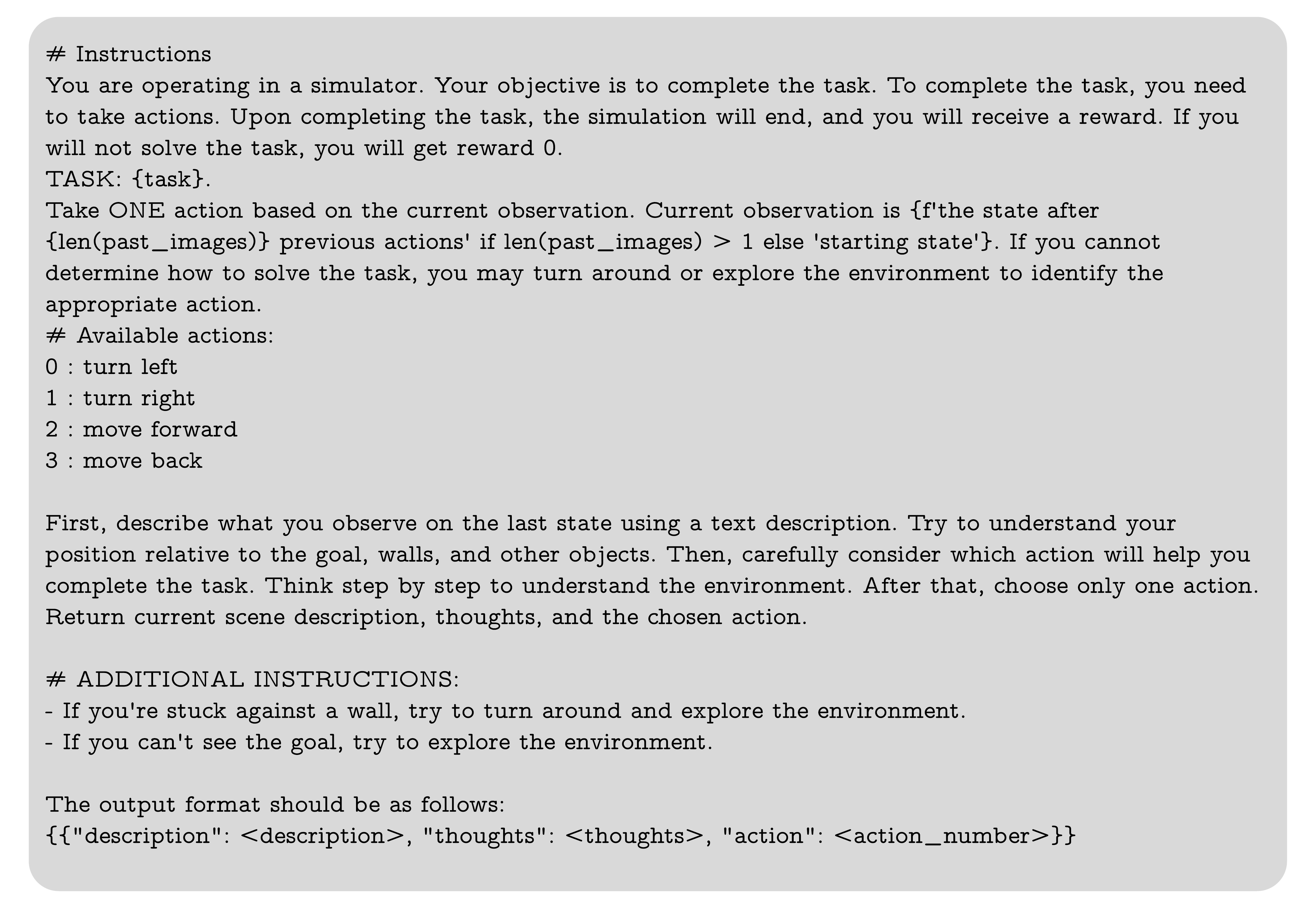}
    \caption{Example of a prompt template for MiniWorld environments.}
    \label{fig:input_example}
\end{figure}

Figure \ref{fig:input_example} shows an illustrative example of a template prompt for our environments.

\section{Appendix B: Hyperparameters}\label{app:hyper}
We list parameters for training our approach in Table \ref{tab:hyperparams_ours}. RL4VLM hyperparameters can be found in Table \ref{tab:hyperparams_rl4vlm}. We aimed to search the same hyperparameter space; however, since RL4VLM requires a wider search due to $\lambda$, the resulting range is somewhat narrower. If a run with a given hyperparameter was completed, then for the final comparison we took the best success rate. For the OneRoom and ALFWorld environments, curves are plotted as mean~$\pm$~std over 4 seeds. Additionally, for the LOOP experiments we searched over the hyperparameters in Table \ref{tab:hyperparams_loop}.

\begin{table}[t]
    \centering
    \resizebox{\columnwidth}{!}{%
    \begin{tabular}{l|c|c|c}
        \toprule
        Hyperparameter & MiniWorld & ALFWorld & EZPoints \\
        \midrule
        Env. Steps & 51200 & 51200 & 51200   \\
        Learning Rate (init $\rightarrow$ final) & 5e-5 $\rightarrow$  1e-7  & 5e-5 $\rightarrow$  1e-7 & 5e-5 $\rightarrow$  1e-7   \\
        Scheduler & cosine & cosine & cosine   \\
        GAE $\lambda_g$ & $\{0.95, 0.99\}$ & $\{0.9, 0.95, 1\}$ & 1  \\
        $\gamma_g$ & $\{0.99, 1\}$ & $\{0.9, 0.99, 1\}$ & 1 \\
        Value Loss Coeff. & 0.15 & 0.15 & 0.15 \\
        KL $\beta$ & 0.05 & 0.05 & $\{0.05, 0.15\}$ \\
        Policy Freeze (steps) & 2 & 2 & 2 \\
        Grad Accum. Steps & 128 & 128 & 128 \\
        Mini-batch Size & 1 & 1 &  1  \\
        PPO Epochs & 2 & 2 & 2 \\
        Eval Episodes & 50 & 50 & 50 \\
        Obs. Image Length & 4 & 4 &  1 \\
        \bottomrule
    \end{tabular}
    }
    \caption{Training hyperparameters per environment for our approach. Values are shared across environments unless specified otherwise.}
    \label{tab:hyperparams_ours}
\end{table}

\begin{table}[t]
    \centering
    \resizebox{\columnwidth}{!}{%
    \begin{tabular}{l|c|c|c}
        \toprule
        Hyperparameter & MiniWorld & ALFWorld & EZPoints \\
        \midrule
        Env. Steps & 51200 & 51200 & 51200   \\
        Learning Rate (init $\rightarrow$ final) & 5e-5 $\rightarrow$  1e-7  & 5e-5 $\rightarrow$  1e-7 & 5e-5 $\rightarrow$  1e-7   \\
        Scheduler & cosine & cosine & cosine   \\
        GAE $\lambda_g$ & 0.95 & $\{0.9, 1\}$ & 1  \\
        $\gamma_g$ & 0.99 & $\{0.9, 1\}$ & 1 \\
        Value Loss Coeff. & 0.15 & 0.15 & 0.15 \\
        KL $\beta$ & 0.05 & 0.05 & $\{0.05, 0.15\}$ \\
        Policy Freeze (steps) & 2 & 2 & 2 \\
        Grad Accum. Steps & 128 & 128 & 128 \\
        Mini-batch Size & 1 & 1 &  1  \\
        PPO Epochs & 2 & 2 & 2 \\
        Eval Episodes & 50 & 50 & 50 \\
        Obs. Image Length & 4 & 4 &  1 \\
        \bottomrule
    \end{tabular}
    }
    \caption{Training hyperparameters per environment for RL4VLM. Values are shared across environments unless specified otherwise.}
    \label{tab:hyperparams_rl4vlm}
\end{table}

\begin{table}[t]
    \centering
    \resizebox{\columnwidth}{!}{%
    \begin{tabular}{l|c}
        \toprule
        Hyperparameter & MiniWorld \\
        \midrule
        Algorithm steps & 75    \\
        Learning Rate (init $\rightarrow$ final) & 5e-5 $\rightarrow$  1e-7   \\
        Scheduler & cosine    \\
        $\gamma_g$ & $\{0.99, 1\}$  \\
        KL $\beta$ & 0.05 \\
        Grad Accum. Steps & 128 \\
        Mini-batch Size & 1   \\
        PPO Epochs & 2 \\
        Eval Episodes & 50  \\
        Obs. Image Length & 4  \\
        \bottomrule
    \end{tabular}
    }
    \caption{Training hyperparameters on MiniWorld for LOOP.}
    \label{tab:hyperparams_loop}
\end{table}
\section{Appendix C: Qwen2-VL-7b Evaluation Setup}\label{app:qwen-eval}
For most benchmarks we use the lmm-eval framework. Since scores from the original Qwen2-VL paper are not easily reproducible, we reran most evaluations. The evaluation hyperparameters are $max\_pixels=200704, min\_pixels=3136, max\_num\_frames=32$. For Balrog and skill-specific benchmarks, we use separate codebases.

\section{Appendix D: Detailed Training Dynamics}\label{app:results_full}
\begin{figure}[th!]
    \centering
    \includegraphics[width=0.45\textwidth]{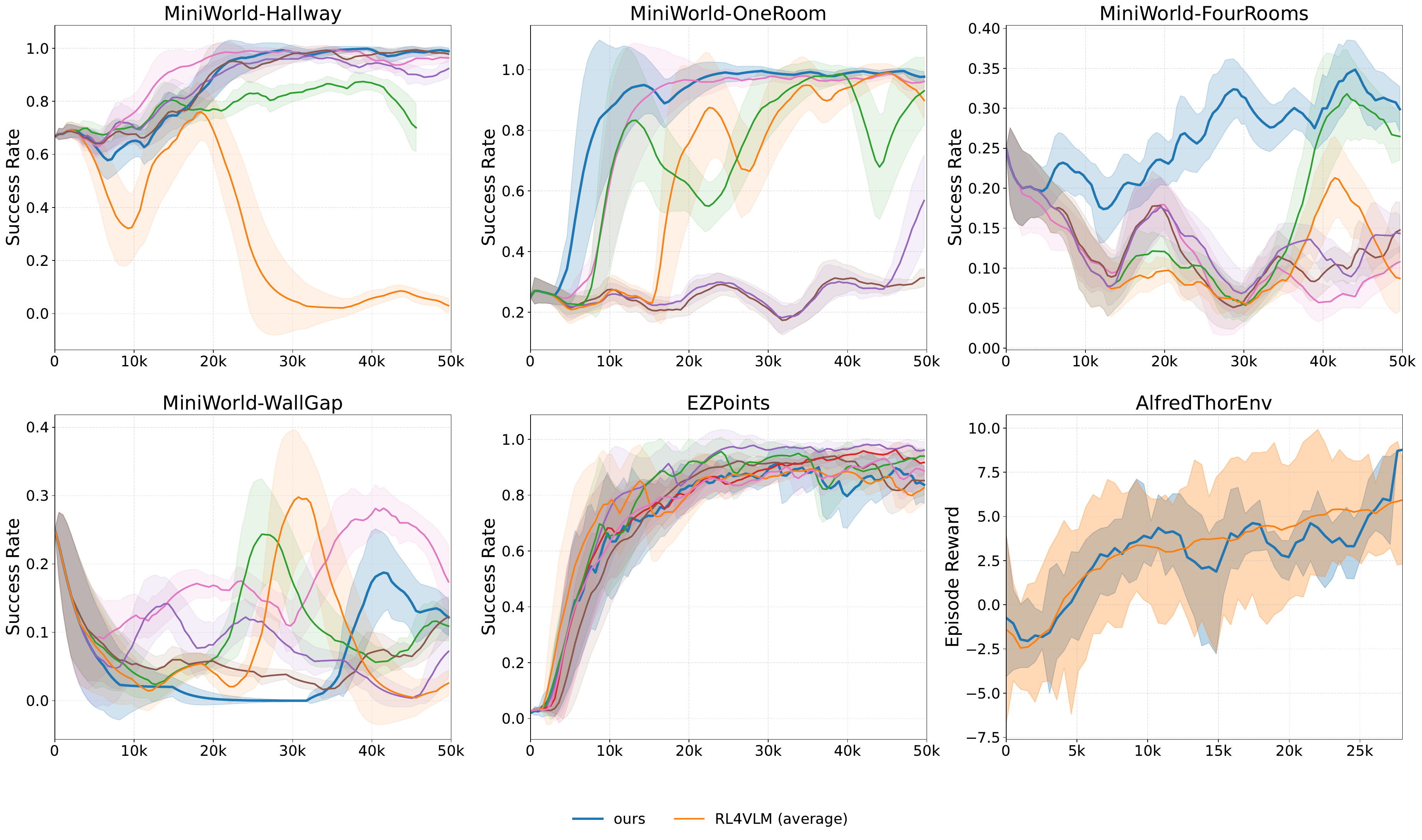}
    \caption{Episode success rates without averaging over the thought-probability coefficient.}
    \label{fig:methods_success_rates_no_average}
\end{figure}

In Figure \ref{fig:methods_success_rates_no_average}, results without averaging over the thought coefficient $\lambda$ can be found.

\section{Appendix E: ArCHer On-policy runs}
\begin{figure}[th!]
    \centering
    \includegraphics[width=0.45\textwidth]{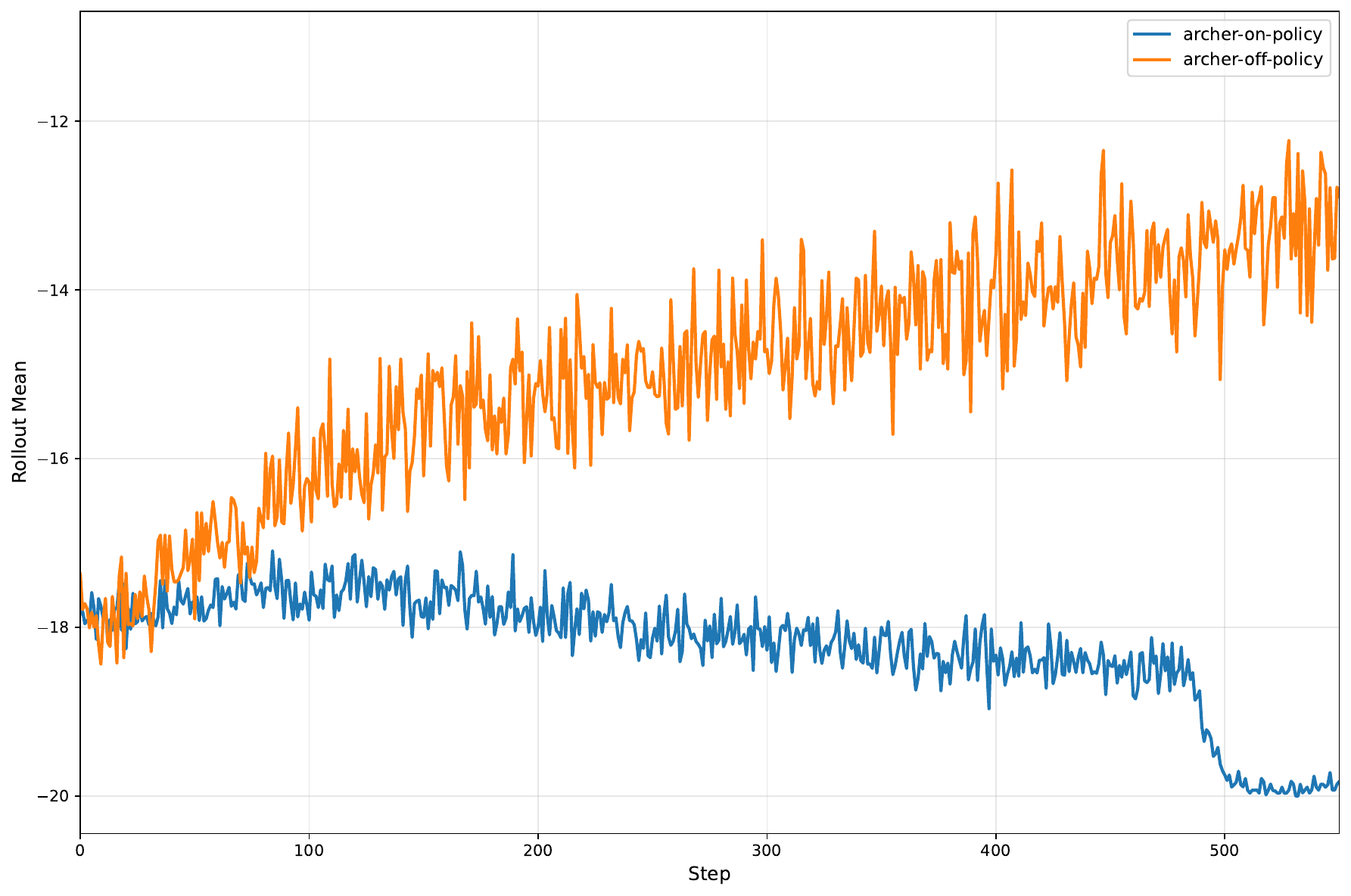}
    \caption{20Q LMRL performance curves for on-policy and off-policy ArCHer.}
    \label{fig:archer}
\end{figure}

In Figure \ref{fig:archer} we show how ArCHer performs differently under off-policy (very large buffer, which is hard to maintain in the case of images and videos) and on-policy (replay buffer equals rollout size) scenarios. In this experiment we use the 20Q environment from LMRL. The rollout size equals 512, while the replay buffer in the off-policy scenario equals 100k. For the on-policy setup, we also experimented with $\tau$ (in Polyak averaging) and plotted the curve for the best value. All other parameters were set to the defaults in the ArCHer paper.

\end{document}